%% file: emnlp2022.tex
\pdfoutput=1

\documentclass[11pt]{article}

\usepackage{EMNLP2022}

\usepackage{times}
\usepackage{latexsym}

\usepackage[T1]{fontenc}

\usepackage[utf8]{inputenc}

\usepackage{microtype}

\usepackage{inconsolata}

\usepackage{booktabs}       %
\usepackage{amsfonts}       %
\usepackage{nicefrac}       %
\usepackage{microtype}      %
\usepackage{xcolor}         %
\usepackage{standalone}
\usepackage{latexsym}
\usepackage{amsmath}
\usepackage{amssymb}
\usepackage{amsthm}
\usepackage{graphicx}
\usepackage{subcaption}
\usepackage{array}
\usepackage{tabu}
\usepackage{makecell}
\usepackage{paralist}
\usepackage{cases}
\usepackage{diagbox}
\usepackage{enumitem}
\usepackage{soul}
\usepackage{multirow}
\usepackage{verbatim}
\usepackage{tabulary}
\usepackage{booktabs}
\usepackage[mathscr]{euscript}
\usepackage{mathtools}
\usepackage{algorithm}
\usepackage{colortbl}
\usepackage{tabstackengine}
\usepackage{tikz}
\usepackage{relsize}
\usepackage{microtype}
\usepackage{tablefootnote}
\usepackage{silence}
\usepackage{bbm}

\usepackage{tikz}
\usepackage{soul}

\definecolor{lightgray}{gray}{0.9}
\colorlet{soulgreen}{green!30}
\definecolor{red}{HTML}{FF0000}
\definecolor{blue}{HTML}{0000FF}
\definecolor{darkgreen}{HTML}{228B22}
\definecolor{dblue}{HTML}{007FFF}

\newcommand{\textred}[1]{\textcolor{red}{#1}}
\newcommand{\textblue}[1]{\textcolor{blue}{#1}}

\usepackage{booktabs,arydshln}

\makeatletter
\def\adl@drawiv#1#2#3{%
        \hskip.5\tabcolsep
        \xleaders#3{#2.5\@tempdimb #1{1}#2.5\@tempdimb}%
                #2\z@ plus1fil minus1fil\relax
        \hskip.5\tabcolsep}
\newcommand{\cdashlinelr}[1]{%
  \noalign{\vskip\aboverulesep
           \global\let\@dashdrawstore\adl@draw
           \global\let\adl@draw\adl@drawiv}
  \cdashline{#1}
  \noalign{\global\let\adl@draw\@dashdrawstore
           \vskip\belowrulesep}}
\makeatother

\usepackage[noend]{algpseudocode}
\algnewcommand{\parState}[1]{\State%
    \parbox[t]{\dimexpr\linewidth-\algmargin}{\strut\hangindent=\algorithmicindent \hangafter=1 #1\strut}}

\algrenewcommand\algorithmicindent{1.0em}%

\usepackage{stmaryrd}
\usepackage{tikz-dependency}
\usetikzlibrary{automata,decorations.markings,arrows,positioning,matrix,calc,patterns,angles,quotes,calc}
\usepackage{adjustbox}
\usepackage{tabularx}
\usepackage{xspace}
\usepackage{tabulary}
\usepackage{afterpage}

\usepackage[a-1b]{pdfx}
\usepackage{hyperref}
\usepackage[utf8]{inputenc}
\usepackage{url}
\usepackage{bm}
\usepackage{color}
\usepackage{graphicx}
\usepackage{slashbox}
\usepackage[toc,page]{appendix}
\usepackage{makecell}
\usepackage{boldline}

\definecolor{orange}{rgb}{1,0.5,0}
\definecolor{mdgreen}{rgb}{0.05,0.6,0.05}
\definecolor{mdblue}{rgb}{0,0,0.7}
\definecolor{dkblue}{rgb}{0,0,0.5}
\definecolor{dkgray}{rgb}{0.3,0.3,0.3}
\definecolor{slate}{rgb}{0.25,0.25,0.4}
\definecolor{gray}{rgb}{0.5,0.5,0.5}
\definecolor{ltgray}{rgb}{0.7,0.7,0.7}
\definecolor{purple}{rgb}{0.7,0,1.0}
\definecolor{lavender}{rgb}{0.65,0.55,1.0}

\definecolor{mypurple}{RGB}{111,61,121}
\definecolor{myblue}{RGB}{46,88,180}
\definecolor{myred}{RGB}{181,68,106}
\definecolor{myyellow}{RGB}{204,143,55}
\definecolor{mygray}{RGB}{128,128,128}
\definecolor{mygreen}{RGB}{126,198,54}

\newcommand{\interalia}[1]{\citep[\emph{inter alia}]{#1}}

\DeclareSymbolFont{extraup}{U}{zavm}{m}{n}
\DeclareMathSymbol{\vardiamond}{\mathalpha}{extraup}{87}

\newcolumntype{L}[1]{>{\raggedright\let\newline\\\arraybackslash\hspace{0pt}}m{#1}}
\newcolumntype{C}[1]{>{\centering\let\newline\\\arraybackslash\hspace{0pt}}m{#1}}
\newcolumntype{R}[1]{>{\raggedleft\let\newline\\\arraybackslash\hspace{0pt}}m{#1}}

\theoremstyle{definition}

\theoremstyle{remark}

\algrenewcommand{\algorithmiccomment}[1]{\leavevmode$\triangleright$ #1}

\setul{1pt}{.4pt}

\usepackage[shortcuts]{extdash}  %

\usepackage{blindtext}
\usepackage{graphicx}
\usepackage{capt-of}
\usepackage{booktabs}
\usepackage{varwidth}
\newsavebox\tmpbox
\input{math_commands.tex}

\newcommand{\resolved}[1]{}

\newcommand{\com}[1]{}
\newcommand{\alt}{\textsc{Twist}\xspace}
\newcommand{\twist}{\textsc{Twist}\xspace}

\newcommand{\bos}{\textsc{bos}\xspace}
\newcommand{\eos}{\textsc{eos}\xspace}

\newcommand{\mapoutput}{\mathrm{map\_output}}

\newcommand{\vocab}{\mathcal{V}}

\usepackage[skins]{tcolorbox}

\usepackage{algorithm}
\usepackage{colortbl}
\usepackage{tabstackengine}
\usepackage{tikz}
\usepackage{relsize}
\usepackage{microtype}
\definecolor{magenta}{HTML}{F3DFF1}
\definecolor{hlgreen}{HTML}{ccfcc4}
\definecolor{figblue}{HTML}{e7f2fe}

\DeclareRobustCommand{\hlfb}[1]{{\sethlcolor{figblue}\hl{#1}}}

\DeclareRobustCommand{\greenbox}[1]{\setlength{\fboxsep}{1.0pt}\colorbox{green!25}{#1}}
\DeclareRobustCommand{\redbox}[1]{\setlength{\fboxsep}{1.0pt}\colorbox{red!15}{#1}}

\newcommand{\thumbup}[0]{\raisebox{-.2\height}{\includegraphics[width=.02\textwidth]{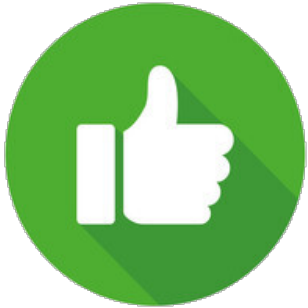}}}
\newcommand{\thumbdown}[0]{\raisebox{-.2\height}{\includegraphics[width=.02\textwidth]{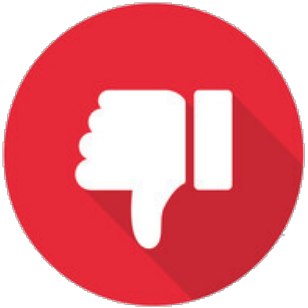}}}

\title{Twist Decoding: Diverse Generators Guide Each Other}

\newcommand*\samethanks[1][\value{footnote}]{\footnotemark[#1]}

\author{
    Jungo Kasai$^{\heartsuit\clubsuit}$
\quad
\textbf{Keisuke Sakaguchi}$^{\diamondsuit}$\thanks{\hspace{2mm}This work was done while Keisuke Sakaguchi was at the Allen Institute for AI and Hao Peng was at the University of Washington.}
\quad
\textbf{Ronan Le Bras}$^{\clubsuit}$
\quad
\textbf{Hao Peng}$^{\clubsuit}$\samethanks
\\
\textbf{Ximing Lu}$^{\heartsuit\clubsuit}$
\quad
\textbf{Dragomir Radev}$^{\spadesuit}$
\quad
\textbf{Yejin Choi}$^{\heartsuit\clubsuit}$
\quad
\textbf{Noah A.\ Smith}$^{\heartsuit\clubsuit}$\\ 
\vspace{-0.3cm}
\\
$^{\heartsuit}$Paul G.\ Allen School of Computer Science \& Engineering, University of Washington
    \\
     $^{\clubsuit}$Allen Institute for AI  \quad
     $^{\diamondsuit}$Tohoku University \\
     $^{\spadesuit}$Department of Computer Science, Yale University \\
    {\tt \{jkasai,lux32,yejin,nasmith\}@cs.washington.edu} \ {\tt keisuke.sakaguchi@tohoku.ac.jp}\\
    {\tt \{ronanlb,haop\}@allenai.org} \quad {\tt dragomir.radev@yale.edu}
}

\begin{document}
\maketitle
\setlength{\abovedisplayskip}{2pt}
\setlength{\belowdisplayskip}{2pt}
\input{text/abstract}
\input{text/intro}
\input{text/method}
\input{text/experiments}
\input{text/analysis}
\input{text/related}
\input{text/conclusion}

\input{text/limitation}

\input{text/acknowledgement}

\bibliography{acl}
\bibliographystyle{acl_natbib}

\clearpage
\appendix
\input{text/appendix}

\end{document}

%% file: math_commands.tex
\usepackage{amsmath,amsfonts,bm}

\def\eqref#1{equation~\ref{#1}}
\def\1{\bm{1}}

\def\rvy{{\mathbf{y}}}
\def\rvz{{\mathbf{z}}}

\def\vy{{\mathbf{y}}}
\def\vz{{\mathbf{z}}}

\DeclareMathAlphabet{\mathsfit}{\encodingdefault}{\sfdefault}{m}{sl}
\SetMathAlphabet{\mathsfit}{bold}{\encodingdefault}{\sfdefault}{bx}{n}

\DeclareMathOperator*{\topk}{topk}

%% file: text/abstract.tex
\begin{abstract}
Many language generation models are now available for a wide range of generation tasks, including machine translation and summarization.
Combining such diverse models may lead to further progress, but ensembling generation models is challenging during inference:
conventional ensembling methods (e.g., shallow fusion) require that the models share vocabulary/tokenization schemes. 
We introduce \twist decoding, a simple and general text generation algorithm that benefits from diverse models at inference time.
Our method does not assume the vocabulary, tokenization or even generation order is shared.
Our extensive evaluations on machine translation and scientific paper summarization demonstrate that \twist decoding substantially outperforms each model decoded in isolation over various scenarios, including cases where domain-specific and general-purpose models are both available.
\twist decoding also consistently outperforms the popular reranking heuristic where output candidates from one model are rescored by another.
We hope that our work will encourage researchers and practitioners to examine generation models collectively, not just independently, and to seek out models with complementary strengths to the currently available models.\footnote{Our code is available at \url{https://github.com/jungokasai/twist_decoding}.}
\end{abstract}

%% file: text/intro.tex
\begin{figure}[t!]
\centering
    \includegraphics[width=0.49\textwidth]{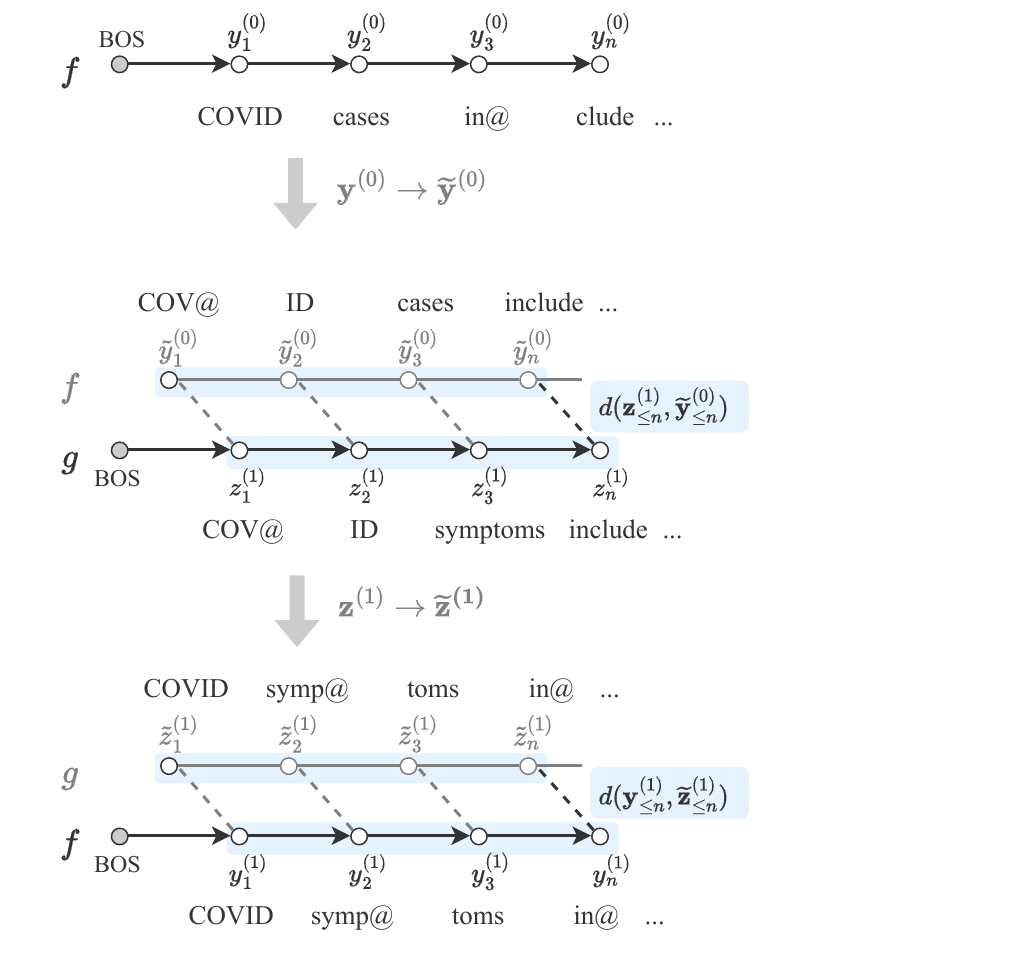}
\caption{
\twist decoding of two generation models, $f$ and $g$, that does not assume a shared vocabulary, tokenization, or generation order.
Beam search is first applied to $f$ to generate $\vy^{(0)}$, followed by output mapping to $\widetilde{\vy}^{(0)}$ (e.g., $f$'s detokenization and $g$'s tokenization or sequence reversal).
$g$ is then decoded with beam search augmented with distances from the set of previously-generated outputs (here only one sequence $\mathbf{y}$ is shown): \hlfb{$d(\vz_{\leq n}^{(1)}, \widetilde{\vy}_{\leq n}^{(0)})$}.
Subsequently, $f$ is similarly decoded with $g$'s guidance.
Here we show one iteration that already achieves substantial improvements (\S\ref{sec:analysis}).
@ indicates the BPE separator.
}
\label{fig:twist_decoding}
\end{figure}

\section{Introduction}
Natural language generation is an important building block for many applications, such as machine translation, summarization, and question answering \interalia{ng-etal-2019-facebook,lewis-etal-2020-bart,T5,gpt3,asai2021cora}.
Researchers have recently explored and advanced models for generation in various aspects, including model architecture \cite{Bahdanau2014NeuralMT,Vaswani2017AttentionIA}, domain adaptation \cite{chu-wang-2018-survey,bapna-firat-2019-simple}, prompting \cite{gpt3}, and even generation order \cite{Gu2017NonAutoregressiveNM}.
The resulting generation models are diverse, trained on different data, with different assumptions, at different times.  We hypothesize that diverse generation models may achieve better results through ensembling, if the various approaches have complementary strengths.  Given the high cost of unifying approaches during training time \cite{strubell-etal-2019-energy,greenai}, inference-time combination of existing models is an attractive alternative.

One well-established ensembling technique is ``shallow fusion'' \interalia{Sutskever2014SequenceTS,shallow-deep-fusion,firat-etal-2016-zero}, which aggregates models' scores during beam search.  This approach requires, however, that the models use the same vocabulary/tokenization scheme and organize the search in the same way (e.g., autoregressive, left-to-right factorization).

We introduce a new inference algorithm, \twist decoding (Fig.\ \ref{fig:twist_decoding}), that enables more diverse generators to guide each other.
\twist decoding can combine generators with different vocabularies, (de)tokenization, and even generation order without any additional training or finetuning.
Our method decodes a model by standard beam search, but the scores at every step incorporate a simple function that measures the distance from outputs of the other model.
We run this procedure on each generation model in turn, so that both can benefit from each other.

We present extensive experiments on machine translation and scientific paper summarization and show that \twist decoding can improve performance over each model decoded in isolation across several scenarios: combining 1) generic and domain-specific models, 2) left-to-right and right-to-left generation models, and 3) models that generate using different conditioning inputs.
Our results show consistent performance gains from combining generic and domain-specific translation models over a wide range of domains, including medical and legal translation.
Applications in these domains require particularly high accuracy, and \twist decoding is a desirable alternative to standard beam search on a single model.
Interestingly, we find that \twist decoding between generic and domain models is effective even when parallel data from the domain are scarce and the domain model yields poor performance by itself, suggesting complementary
strengths of diverse generators (\S\ref{sec:low-resource}).

\twist decoding can be seen as a generalization of reranking heuristics that have proven effective in syntactic parsing \cite{shen-joshi-2003-svm,charniak-johnson-2005-coarse,collins-koo-2005-discriminative}, speech recognition \cite{collins-etal-2005-discriminative}, and machine translation \cite{shen-etal-2004-discriminative,och-etal-2004-smorgasbord}: one model generates candidate sequences, followed by rescoring from another model.
We present extensive comparisons with reranking baselines and demonstrate that \twist decoding outperforms reranking consistently.
We also observe that since the encoder computations on two models can be parallelized, the inference time required for \twist decoding is much shorter than the sum of the two models, resulting only in a 50\% increase, relative to decoding of a single model in isolation (\S\ref{sec:analysis}).
\twist decoding is therefore a viable alternative to standard beam search on a single model and the widespread reranking heuristic.

%% file: text/method.tex
\begin{figure}[t!]
\algrenewcommand{\algorithmiccomment}[1]{\hskip1em\# \footnotesize#1}
\begin{algorithm}[H]
\small
$k$: beam size. \ $M$: maximum length.\\
$\vocab_g$: vocabulary of $g$. \ $g(\cdot)$: scoring function.\\
$\mathcal{Y}^{(t-1)}$: set of output sequences from $f$. \ $\mathcal{Z}^{(t)}$: new outputs.\\
$B_n$: beam of continuing sequences.\\
$H$: expanded hypotheses before beam selection.\\
\greenbox{$d(\cdot,\cdot)$}: distance between partial sequences.\\
\greenbox{$\lambda_f$}: scalar coefficient for the distance.
\begin{algorithmic}[1]
\small
\State $\widetilde{\mathcal{Y}}^{(t-1)} =  \left\{ \widetilde{\vy} = \mapoutput(\vy) \mid \vy \in \mathcal{Y}^{(t-1)} \right\}$ \label{line:map_output}
\small
\State $B_{0} \gets \{ \bos \}$, \ $\mathcal{Z}^{(t)}  \gets \varnothing $ 
\small
\For{ $n \in \{ 1, \dots, M \}$ }
\small
    \State $H \gets \varnothing$
\small
    \For{$ \vz \in B_{n-1}$} \algorithmiccomment{Expansion.}
\small
\small
    \For{$z \in \vocab_g$}
\small
        \State $s \gets g(\rvz \circ z)$\colorbox{green!25}{$- \lambda_f \min_{\widetilde{\vy} \in \widetilde{\mathcal{Y}}^{(t-1)}} d \left(\rvz \circ z,  \widetilde{\rvy}_{\leq n}\right)$}  \label{line:bs-score-eval}
        \label{line:distance}
        \State $H.\mathrm{add}( \langle s, \vz \circ z \rangle)$
\small
    \EndFor
    \EndFor
\small
\State $B_n \gets \topk(H)$, \ $\mathcal{Z}^{(t)}.\mathrm{add}\left(\mathrm{finished}(H)\right)$
\small
\EndFor
\small
\State \Return $\mathcal{Z}^{(t)}$

\end{algorithmic}
\caption*{\small \textbf{\twist Decoding}\\$g$ generates $\vz$ with guidance from $f$ at iteration $t$}
\end{algorithm}
\caption{
\twist decoding when $g$ is guided by $f$.
Swap $f$ and $g$ and $\vy$ and $\vz$ to obtain $\mathcal{Y}^{(t)}$.
$\mapoutput$ converts outputs from $f$ to $g$; e.g., $f$'s detokenization followed by $g$'s tokenization. It would also include sequence reversal if $f$ or $g$ is a right-to-left model.
The \greenbox{highlighted line} is the \textit{only} modification that \twist decoding introduces to standard beam search.
The input sequence to $g$ is omitted.
See also \citet{kasai2022beampatience} for the stopping criterion and implementation details (the \textit{first come, first served} heuristic).
}
\label{alg:twist}
\end{figure}
\section{\alt Decoding}
\label{sec:twist_decoding}
We propose \twist decoding, a general decoding algorithm that generates text from diverse models without assumptions of a shared vocabulary, tokenization, or generation order.
At the core of the algorithm is a simple modification in standard beam search (\greenbox{highlighted} in Fig.\ \ref{alg:twist}); we incorporate into a scoring function the distance from outputs that are previously generated by another model.

\subsection{Initial Decoding}
Let us assume that we have two generation models: $f$ and $g$.\footnote{The algorithm can be readily extended to three or more generators. We also abuse $f$ or $g$ to mean both the generator and its scoring function.}
Both $f$ and $g$ assign scores to output sequences.\footnote{They typically assign log-probabilities, but it is not necessary to assume the scores form a valid probability distribution.}
For example, $f$ can be a domain-specific translation model and $g$ a generic one.
$f$ and $g$ perform their own pre/postprocessing (e.g., (de)tokenization) and factorization (e.g., left-to-right or right-to-left factorization).
Here we suppress for brevity the conditional dependence on the input (e.g., machine translation input from the source language).
 Standard beam search with a beam size $k$ is first applied to $f$ to produce a set of $k$ output sequences: $\mathcal{Y}^{(0)}$.
 This approximately solves $\topk_{\vy} f\left( \vy \right)$ by pruned breadth-first search, and often returns higher-quality outputs than the exact search counterpart \cite{stahlberg-byrne-2019-nmt,meister-etal-2020-beam}.

\subsection{Mutually-Guided Decoding}
\label{sec:mutual_guidance}
Once $\mathcal{Y}^{(0)}$ is obtained, we proceed with decoding generators with mutual guidance (Fig.\ \ref{alg:twist}; $t \geq 1$).
\paragraph{Output Sequence Mapping.}
The commonly-used technique of ensembling \cite{Sutskever2014SequenceTS} or shallow fusion \cite{shallow-deep-fusion,stahlberg-etal-2018-simple} adds scores from $f$ and $g$ at every step and executes the same search algorithm to approximately solve $\topk_{\vy} f(\vy) + g(\vy)$.
This method thus necessitates a shared vocabulary, tokenization, and generation order \cite{imamura-sumita-2017-ensemble}.
We relax this assumption and first map the candidates in $\mathcal{Y}^{(t-1)}$ to output sequences for $g$: $\widetilde{\mathcal{Y}}^{(t-1)}$ (Line \ref{line:map_output} in Fig.\ \ref{alg:twist}).
This mapping ($\mapoutput$) typically involves deterministic operations of $f$'s detokenization followed by $g$'s tokenization.
Sequence reversal is also performed if $f$ and $g$ generate in the opposite order.
For example, if $g$ uses byte-pair encoding \cite{sennrich-etal-2016-neural}, but $f$ does not, we might have $\vy\!=$\textit{John does n't like Mary} mapped to $\widetilde{\vy}\!=$\textit{Jo@ hn doesn't like Mar@ y}, where @ denotes subword separation.

\paragraph{Decoding with Distance Terms.}
We then decode $g$ with guidance from $\widetilde{\mathcal{Y}}^{(t-1)}$.
Specifically, we perform beam search with a simple modification in scoring (\greenbox{Line \ref{line:distance}}).
In this work, we use a simple distance measure that adds binary distances at all positions (i.e., the Hamming distance):
\begin{align*}
d \left ( \vz_{\leq n}, \widetilde{\vy}_{\leq n} \right) =  \sum_{i \leq n} \mathbbm{1} \left\{z_{i} \neq \widetilde{y}_i \right\}
\end{align*}
We also explored using  the distance between (sub)word embeddings from the model: $\sum_{i\leq n}\lVert e (z_i) - e (\widetilde{y}_i) \rVert_2$, but this did not bring improvements (\S\ref{sec:analysis}).
Note also that when $i$ exceeds the length of $\widetilde{\vy}$, we assume $\widetilde{y}_i = \eos$.
The overall distance term is then 
\begin{align*}
\min_{\widetilde{\vy} \in \widetilde{\mathcal{Y}}^{(t-1)}} d \left ( \vz_{\leq n}, \widetilde{\vy}_{\leq n} \right)
\end{align*}
Here we minimize over the output sequences to compute the distance to the closest candidate.
These candidates from $\widetilde{\mathcal{Y}}^{(t-1)}$ can be equally good outputs but differ only by one word; in such cases, this minimization operation avoids overestimation of the distances.
The new score at step $n$ in beam search is now computed by:
\begin{align*}
 g(\vz_{\leq n}) - \lambda_f \min_{\widetilde{\vy} \in \widetilde{\mathcal{Y}}^{(t-1)}} d \left(\vz_{\leq n},  \widetilde{\rvy}_{\leq n}\right),
\end{align*}
where $\lambda_f$ is a scalar coefficient for the distance term that controls the importance of $f$ relative to $g$.
We tune $\lambda_f \in \{0.1, 0.3, 1.0, 3.0\}$ during development.
After this beam search, we obtain a new candidate set, $\mathcal{Z}^{(t)}$.
We then run the same beam search (Fig.\ \ref{alg:twist}) with the roles of $f$, $\mathcal{Y}$ and $g$, $\mathcal{Z}$ swapped.\footnote{We can stop inference with $\mathcal{Z}^{(t)}$, but we found that led to performance degradation in preliminary development.}
Namely, we decode $f$ with distance terms from $\mathcal{Z}^{(t)}$ at each step of beam search:
\begin{align*}
 f(\vy_{\leq n}) - \lambda_g \min_{\widetilde{\vz} \in \widetilde{\mathcal{Z}}^{(t)}} d \left(\vy_{\leq n},  \widetilde{\vz}_{\leq n}\right)
\end{align*}
Finally, the highest-scoring sequence from $\mathcal{Y}^{(t)}$ is output.
This process of mutually-guided decoding can be repeated multiple times.
We observe, however, that one iteration ($t\!=\!1$) suffices to bring performance gains (\S\ref{sec:analysis}).
We also present detailed sensitivity analysis over varying $\lambda_f$ and $\lambda_g$ and find that \twist decoding is particularly effective when $\lambda_g > \lambda_f$ (i.e., initial exploration by $g$ is encouraged with relatively little guidance from $f$'s original outputs; see \S\ref{sec:analysis}).

\paragraph{Reranking Heuristic as a Special Case.}
Notice that as $\lambda_{f} \rightarrow \infty$, $g$'s generation falls back to a reranking heuristic: top $k$ sequences from the initial $f$ decoding are reranked according to $g$.
This reranking heuristic has proven successful in a wide range of sequence generation tasks, including machine translation \cite{shen-etal-2004-discriminative}, syntactic parsing \cite{collins-koo-2005-discriminative}, and speech recognition \cite{collins-etal-2005-discriminative}.
Reranking is performed in many strong machine translation systems to use a right-to-left model to improve a left-to-right model; e.g., top-performing systems in recent WMT competitions \cite{ng-etal-2019-facebook,kiyono-etal-2020-tohoku,wang-etal-2021-tencent,akhbardeh-etal-2021-findings}.
In our experiments, we extensively compare performances of \twist decoding and reranking and demonstrate that the former consistently outperforms the latter.

%% file: text/experiments.tex
\section{Experiments}
\label{sec:experiments}
We present experiments across three scenarios: combining domain and generic models for machine translation (\S\ref{sec:domain_generic}), left-to-right and right-to-left machine translation models (\S\ref{sec:l2r_r2l}), and scientific paper summarization models that take as input different parts of the paper (\S\ref{sec:tldr}).
We empirically compare \twist decoding with decoding in isolation and the widely-adopted reranking baselines, illustrating that \twist decoding offers performance improvements in various situations \textit{without} any change to the trained models.

\subsection{Domain and Generic Models}
\label{sec:domain_generic}
Machine translation has now been used for many domains, ranging from everyday conversations to medical documents.
Machine translation models are often trained on large amounts of parallel data, such as the Europarl corpus \cite{koehn-2005-europarl} and the OPUS data \cite{tiedemann-2012-parallel}.
Applying these models to out-of-domain data remains a challenge \cite{koehn-knowles-2017-six,chu-wang-2018-survey}, and users for some of these domains require high accuracy in translation (e.g., medical and legal documents).
We will demonstrate that \twist decoding between general-purpose and domain-specific models is a viable approach to tackle this problem.

\paragraph{Setups.}
We use machine translation datasets over diverse domains from prior work \cite{koehn-knowles-2017-six,hu-etal-2019-domain}: German$\rightarrow$English over medical (1.1M training sentence pairs), legal (720K pairs), Koran (religious text, 480K pairs), and subtitles (14M pairs) domains.\footnote{We excluded the IT domain because we found significant overlap between training and dev./test data.}
For the domain-specific models, we train a base-sized transformer model \cite{Vaswani2017AttentionIA} with a 6-layer encoder and a 6-layer decoder on the training data of each domain.
The top-performing German$\rightarrow$English system from WMT19 \cite{barrault-etal-2019-findings,ng-etal-2019-facebook}\footnote{\url{https://github.com/pytorch/fairseq/tree/main/examples/wmt19}.} is used as the generic model.
This generic model is a large-sized transformer trained on a concatenation of publicly available parallel data, including the Europarl \cite{koehn-2005-europarl} and UN \cite{ziemski-etal-2016-united} corpora with the backtranslation technique \cite{sennrich-etal-2016-improving}.
We follow (de)tokenization \cite{koehn-etal-2007-moses} and byte-pair encoding \cite{sennrich-etal-2016-neural} of previous work \cite{koehn-knowles-2017-six,hu-etal-2019-domain}.\footnote{\url{https://github.com/JunjieHu/dali}.}

For every domain, we evaluate a total of six configurations: decoding of the generic and domain models each in isolation; the reranking baseline and \twist decoding with $f$ being the generic model and $g$ being the domain model, as well as the versions where $f$ and $g$ are swapped to see the effect of the two roles.
In all cases, we use beam size 5 \cite{freitag-al-onaizan-2017-beam} and length penalty 1 \cite{gnmt} and conduct all experiments using the \texttt{fairseq} library \cite{ott-etal-2019-fairseq}.
All performance is measured with the COMET score \cite{rei-etal-2020-comet,comet-wmt} and the \textsc{SacreBLEU} implementation \cite{post-2018-call} of the BLEU score \cite{Papineni2001BleuAM}.
Note that COMET is based on crosslingual contextual representations \cite{conneau-etal-2020-unsupervised}, and recent work showed that it achieves significantly higher correlation with expert human judgment than BLEU and other n-gram-based metrics \cite{billboard,kasai2021thumb}.
More experimental details are described in Appendix \S\ref{appendix:domain_mt}.

\begin{table*}[t]
\addtolength{\tabcolsep}{-1.0pt}
\centering
\small
\begin{tabular}{@{} lcc   m{0.01em}  cc   m{0.01em}   cc m{0.01em}  cc m{0.01em}   cc @{}}
\toprule[.1em]

 \multicolumn{3}{c}{\textbf{German$\rightarrow$English}}
&& \multicolumn{2}{c}{\textbf{Medicine}}
&& \multicolumn{2}{c}{\textbf{Law}}
&& \multicolumn{2}{c}{\textbf{Koran}}
&& \multicolumn{2}{c}{\textbf{Subtitles}}
\\
\cmidrule(lr){5-6}
\cmidrule(lr){8-9}
\cmidrule(lr){11-12}
\cmidrule(lr){14-15}
\textbf{Method} 
& $f$ 
& $g$
&& COMET & BLEU
&& COMET & BLEU
&& COMET & BLEU
&& COMET & BLEU
\\

\midrule[.1em]

Isolation
& Generic
& --
&& 44.5    
& 41.2 
&& 30.6 
&  34.8 
&&  14.4
& 16.6 
&& 34.4 
&31.3
\\
Isolation
& Domain
& --
&& 80.7
& 48.3 
&& 60.7 
&  40.9
&&  8.7
& 17.0
&& 32.3
& 29.0 
\\

\midrule[.05em]

Rerank
& Generic
& Domain
&& 59.6
& 43.5 
&& 56.4
& 36.1
&& 14.7 
& 17.0 
&& 40.3 
&  \textbf{32.3}

\\ 

\rowcolor[gray]{.93}
\textbf{\alt}
& Generic
& Domain
&& 71.6 
& 47.5
&&  61.4
& 40.2 
&& \textbf{16.5}
&  18.5
&& \textbf{41.0}
&  32.2

\\
\multicolumn{3}{l}{$\Delta$ (\textbf{\alt}$-$ Rerank)}
&& \textblue{$+12.0$}
& \textblue{$+4.0$}
&& \textblue{$+5.0$}
& \textblue{$+4.1$}
&& \textblue{$+1.8$}
&  \textblue{$+1.5$}
&& \textblue{$+0.7$}
& \textred{$-0.1$}

\\
 \cdashlinelr{1-15}

Rerank
& Domain
& Generic
&& 75.8
& 48.7
&& 59.9
& 40.8
&& 12.3
&  18.1
&& 36.5 
& 30.3 
\\ 

\rowcolor[gray]{.93}
\textbf{\alt}
& Domain
& Generic
&& \textbf{81.6}
& \textbf{50.1} 
&& \textbf{61.6} 
& \textbf{41.3}
&& 15.3
& \textbf{18.7} 
&&  37.3
&  31.0
\\

\multicolumn{3}{l}{$\Delta$ (\textbf{\alt}$-$ Rerank)}
&& \textblue{$+5.8$}
& \textblue{$+1.4$}
&& \textblue{$+1.7$}
& \textblue{$+0.5$}
&& \textblue{$+3.0$}
&  \textblue{$+0.6$}
&& \textblue{$+0.8$}
& \textblue{$+0.7$}

\\

\bottomrule[.1em]
\end{tabular}
\caption{
Combination of generic and domain-specific translation models.
The generic model is the top-performing translation model in WMT19 \citep{ng-etal-2019-facebook} that is trained on a collection of parallel corpora, such as the Europarl and the UN corpora.
Two settings are considered for the reranking baseline and \twist decoding: $f$ is the generic model and $g$ is the domain model or the reverse.
The best scores are in \textbf{bold}. COMET \cite{rei-etal-2020-comet,comet-wmt} uses crosslingual contextual representations \cite{conneau-etal-2020-unsupervised} and achieves significantly higher correlation with expert human judgment than BLEU \cite{Papineni2001BleuAM} and other alternative metrics \cite{billboard,kasai2021thumb}.
}
\label{tab:domain_results}
\end{table*}%

\paragraph{Results.}
Seen in Table \ref{tab:domain_results} are the results from our experiments over various domains.
Firstly, given two translation models $f$ and $g$, \twist decoding outperforms the reranking baseline in all configurations (indicated in \textblue{blue}) with only one exception (a small drop in BLEU in the subtitles domain). 
Particularly noteworthy are the gains in the medical domain: \twist decoding outperforms the reranking heuristic by \textblue{5.8} COMET and \textblue{1.4} BLEU points when $f$ is the domain model and $g$ is the generic model.
\twist decoding is thus an effective generalization over the reranking heuristic commonly used in the literature across domains.

Comparing the performance of decoding in isolation and \twist decoding, we observe that the best score from \twist decoding substantially outperforms each individual model over all domains: e.g., 81.6 vs.\ 80.7 (domain model) and 81.6 vs.\ 44.5 (generic model) COMET points in the medical domain.
In both medical and legal domains, the generic model underperforms the domain model by a large margin.
Nonetheless, \twist decoding between the two improves over the domain model, suggesting that \twist decoding makes use of their complementary strengths.
Finally, we see a consistent pattern regarding $f$ and $g$: both \twist decoding and the reranking baseline perform better when the higher-performing model is chosen as $f$. (e.g., the domain model performs better in medicine and law, and vice versa in subtitles.)
This is expected because $f$ is used both for initial decoding and final decoding with $g$'s guidance (Fig.\ \ref{fig:twist_decoding}).

\subsection{Left-to-Right and Right-to-Left Models}
\label{sec:l2r_r2l}
Language generation models usually factorize sequences autoregressively in a left-to-right order, but previous work showed that left-to-right (L2R) models can be improved by reranking their outputs with a separate right-to-left (R2L) model \interalia{imamura-sumita-2017-ensemble,ng-etal-2019-facebook,kiyono-etal-2020-tohoku}.
\twist decoding can be readily applied to such scenarios since it does not assume shared generation order between models.

\paragraph{Setups.}
We experiment with two language pairs from the WMT 2020 news translation task \cite{barrault-etal-2020-findings}: Chinese$\rightarrow$English (\textbf{WMT20 ZH-EN}, 48M training sentence pairs) and English$\rightarrow$German (\textbf{WMT20 EN-DE}, 48M pairs).
Submissions for these language pairs to the shared task have human evaluations from professional translators \cite{freitag2021experts}, and the correlation between automatic metrics and the human ratings are studied in subsequent work \cite{billboard}; COMET \cite{comet-wmt,rei-etal-2020-comet} achieves the highest correlation out of the 15+ metrics.

Similar to the previous experiments, we measure all performance using COMET and BLEU scores.
Note that we use two reference translations per instance for WMT20 ZH-EN and three for WMT20 EN-DE, following \citet{billboard}.
They both have reference translations from two different services, and WMT20 EN-DE has an additional translation created by linguists who are asked to paraphrase the two translations as much as possible.
These paraphrased translations are shown to increase correlation with human judgments by mitigating the translationese effect \cite{Graham2020TranslationeseIM} and diversifying the reference \cite{freitag-bleu-paraphrase-references-2020}.
On each dataset, we follow the preprocessing and tokenization \cite{koehn-etal-2007-moses,sennrich-etal-2016-neural} from \citet{billboard}\footnote{\url{https://github.com/jungokasai/billboard/tree/master/baselines}.} and train a large-sized transformer model for left-to-right and right-to-left translation, in which the output English/German sequences are reversed after tokenization.
We implement all models and decoding with \texttt{fairseq} and apply beam search with beam size 5 and length penalty 1.
We again consider a total of six settings: reranking and \twist decoding with L2R as $f$ and R2L as $g$ or the reverse, as well as the individual models.
Further details can be found in Appendix \S\ref{appendix:l2r_r2l}.

\begin{table}[h]
 \addtolength{\tabcolsep}{-3.7pt}  
\centering
\small
\begin{tabular}{@{} l cc @{} cc @{} m{1em} @{} cc  @{}}
\toprule[.1em]

\multicolumn{3}{c}{\textbf{WMT20 Test}}
& \multicolumn{2}{c}{\textbf{ZH$\rightarrow$EN}}
&& \multicolumn{2}{c}{\textbf{EN$\rightarrow$DE}}\\
\cmidrule(lr){4-5}
\cmidrule(lr){7-8}
\textbf{Method}
& $f$
& $g$
& \textsc{COMET} & \textsc{BLEU}
&& \textsc{COMET} & \textsc{BLEU}
\\

\midrule[.1em]
Isolation
&L2R 
& --
&40.8
&35.5
&&42.9
&45.5
\\
Isolation
&R2L
& --
& 40.4
& 35.0
&&
43.3
& 44.9
\\
\midrule[.05em]
Rerank
&L2R
&R2L
& 41.4
& 36.1
&&43.7
& 46.0
\\

\rowcolor[gray]{.93}
\textbf{\alt}
&L2R
&R2L
& 42.8 
& \textbf{36.8}
&& \textbf{45.4}
& \textbf{46.7}
\\

\multicolumn{3}{l}{$\Delta$ (\textbf{\alt}$-$ Rerank)}
& \textblue{$+1.4$}
& \textblue{$+0.7$}
&& \textblue{$+1.7$}
& \textblue{$+0.7$}

\\

\cdashlinelr{1-8}
 
Rerank
& R2L
& L2R
& 41.2
& 35.4
&& 44.7
& 45.2
\\

\rowcolor[gray]{.93}
\textbf{\alt}
& R2L
& L2R
& \textbf{43.1}
& \textbf{36.8}
&& 44.8
& 46.0
\\

\multicolumn{3}{l}{$\Delta$ (\textbf{\alt}$-$ Rerank)}
& \textblue{$+1.9$}
& \textblue{$+1.4$}
&& \textblue{$+0.1$}
& \textblue{$+0.8$}

\\

\bottomrule[.1em]
\end{tabular}
\caption{
Combination of left-to-right (L2R) and right-to-left (R2L) transformer translation models.
ZH: Chinese. DE: German.
Two settings are considered for reranking and our \twist decoding each: L2R or R2L as $f$.
The best scores are in \textbf{bold}.
} 
\label{tab:l2r_r2l}
\end{table}
\paragraph{Results.}
Table \ref{tab:l2r_r2l} shows the results from L2R and R2L translation models.
\twist decoding again outperforms the reranking counterpart by a considerable margin in COMET and BLEU on both language pairs; e.g., 43.1 vs.\ 41.2 COMET points on WMT20 ZH-EN when $f$ is R2L and $g$ is L2R.
The best performance is achieved by \twist decoding on both datasets and improves over the individual models by more than 1 BLEU point.
The reranking baseline, on the other hand, does not outperform L2R in BLEU when $f$ is R2L: 35.4 vs.\ 35.5 (ZH-EN) and 45.2 vs.\ 45.5 (EN-DE).
This result illustrates that \twist decoding is a more effective approach to combine models with different generation order than the popular reranking.

\subsection{Summarization with Different Input}
\label{sec:tldr}
We also experiment with strong models on a highly abstractive scientific paper summarization task: \textbf{SciTLDR} \cite{cachola-etal-2020-tldr}.
Specifically, we use two BART-based models from prior work \cite{cachola-etal-2020-tldr} that differ in input type: one that only takes as input the paper abstract (Abst.) and the other a concatenation of the abstract, introduction, and conclusion (AIC).\footnote{\url{https://github.com/allenai/scitldr}.} 

\paragraph{Setups.}
We use the train/dev./test split from \citet{cachola-etal-2020-tldr}.
Again following \citet{cachola-etal-2020-tldr}, we use all human-written summaries (written either by authors or undergraduate computer science students) as the reference and evaluate performance in terms of the ROUGE score \cite{Lin2004ROUGEAP}.\footnote{We release our models and their outputs, so other metrics can be readily used as well in the future.}
We average the instance-level scores from the Python rouge-score implementation.\footnote{\url{https://pypi.org/project/rouge-score/}.}
Similar to our previous experiments, we use beam size 5 and length penalty 1.
See more detail in Appendix \ref{appendix:tldr}.

\paragraph{Results.}
Table \ref{tab:tldr} presents our results.
\twist decoding substantially outperforms the reranking baseline when $f$ is the AIC model (e.g., \textblue{$+0.5$} ROUGE-L points), but they yield (almost) the same performance when $f$ is the Abst.\ model.
Nonetheless, \twist decoding achieves the best performance out of all configurations.
Our small improvements might be attributed to the fact that the input to the Abst.\ model is a strict subset of the AIC model and there are only limited benefits from combining them.

\begin{table}[h]
 \addtolength{\tabcolsep}{1.5pt}  
\centering
\small
\begin{tabular}{@{} l cc m{0.3mm} @{} ccc @{}}
\toprule[.1em]

\multicolumn{3}{c}{\textbf{SciTLDR Summ.\ Test}}
&& \multicolumn{3}{c}{\textbf{ROUGE}}
\\
\cmidrule(lr){5-7}
\textbf{Method}
& $f$
& $g$
&& \textsc{R-1} & \textsc{R-2} & \textsc{R-L}
\\

\midrule[.1em]
Isolation
&Abst.\
& --
&& 39.9
& 21.1 
& 34.5 
\\
Isolation
&AIC
& --
&& 40.2 
& 21.3
& 34.9 
\\
\midrule[.05em]
Rerank
&Abst.\
&AIC
&& 40.5 
& 21.7
& 35.1
\\

\rowcolor[gray]{.93}
\textbf{\alt}
&Abst.\
&AIC
&& 40.5
& 21.7
&  35.0
\\

\multicolumn{3}{l}{$\Delta$ (\textbf{\alt}$-$ Rerank)}
&& $0.0$
& $0.0$
& \textred{$-0.1$}

\\

\cdashlinelr{1-7}
 
Rerank
& AIC
& Abst.\
&& 40.1 
& 21.2
& 34.8
\\
\rowcolor[gray]{.93}
\textbf{\alt}
& AIC
& Abst.\
&& \textbf{40.7}
& \textbf{22.1}
& \textbf{35.3}
\\

\multicolumn{3}{l}{$\Delta$ (\textbf{\alt}$-$ Rerank)}
&& \textblue{$+0.6$}
& \textblue{$+0.9$}
& \textblue{$+0.5$}
\\
\bottomrule[.1em]
\end{tabular}
\caption{
Combination of scientific paper summarization models.
Both models are BART-based models from prior work \cite{cachola-etal-2020-tldr} with different input: abstract only (Abst.) or abstract, introduction, and conclusion (AIC).
The best scores are in \textbf{bold}.
The ROUGE scores (R-1, R-2, and R-L) are computed by averaging instance-level scores from the Python rouge-score implementation.
} 
\label{tab:tldr}
\end{table}

\subsection{Low-Resource Scenarios}
\label{sec:low-resource}
In our experiments over four diverse domains (\S\ref{sec:domain_generic}), we assumed that plenty of parallel data is available in every domain, and the domain model generally outperformed the generic model.
Concretely, we used 1.1M and 720K training sentence pairs for the medical and legal domains, based on the data splits from previous work \cite{koehn-knowles-2017-six,hu-etal-2019-domain}.
In real-world applications, however, these domain-specific translation data are often scarce since they need to be annotated by bilingual speakers with expertise in those domains.
The question arises: \textit{can a domain model trained on small parallel data still help the generic model by its complementary strengths?}
To simulate such low-resource scenarios, we randomly sample \{10k, 20k, 40k, 80k\} sentence pairs and conduct the same evaluations with the generic and domain models as $f$ and $g$, respectively.

\begin{figure}[h!]
\centering
    \includegraphics[width=0.4\textwidth]{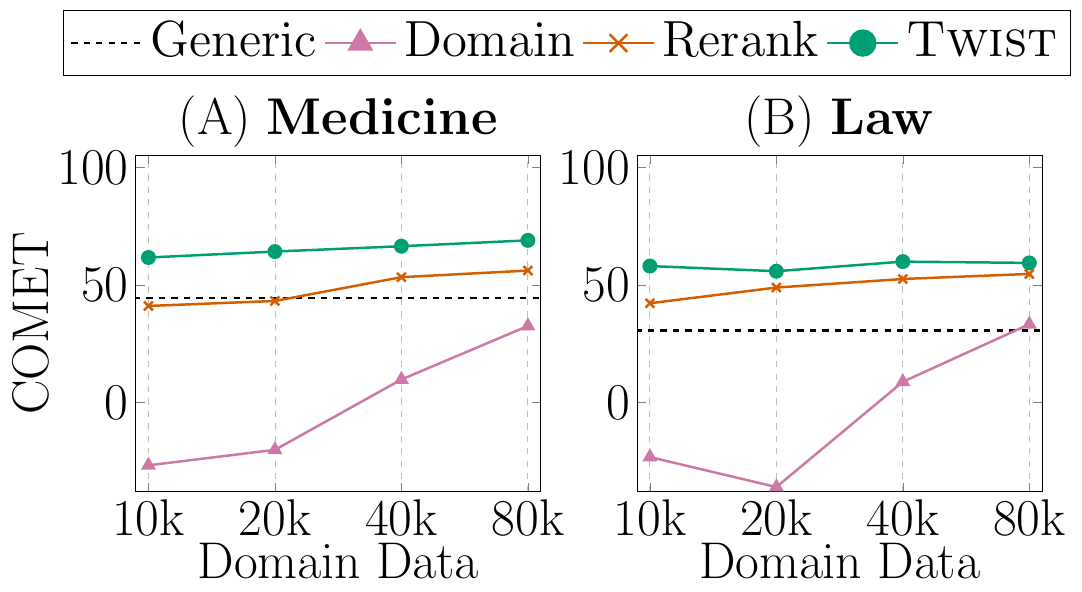}
\caption{Results when parallel data are scarce in the target domain. Both \twist decoding and reranking use the generic model as $f$ and the domain model as $g$. COMET \cite{rei-etal-2020-comet} is a regression-based metric that can take negative values.
$\lambda$s are tuned in each case, and we found that as the domain model ($g$) gets stronger, $\lambda_g$ increases, relative to $\lambda_f$.
This observation is aligned with the intuition that $\lambda_g$ indicates the relative importance of g’s guidance. 
}
\label{fig:low_resource}
\end{figure}

Fig.\ \ref{fig:low_resource} plots COMET scores of various decoding methods on the medical and legal domains.
The score from the generic model is constant because we only change the domain training data.
There is a striking trend: even though the domain model performs poorly by itself, it improves the generic model through \twist decoding over varying sizes.
Reranking also helps the generic model as the data size increases, but the improvement is less pronounced than that of \twist decoding.

%% file: text/analysis.tex
\section{Analysis}
\label{sec:analysis}
\begin{figure}[h!]
\centering \includegraphics[width=0.49\textwidth]{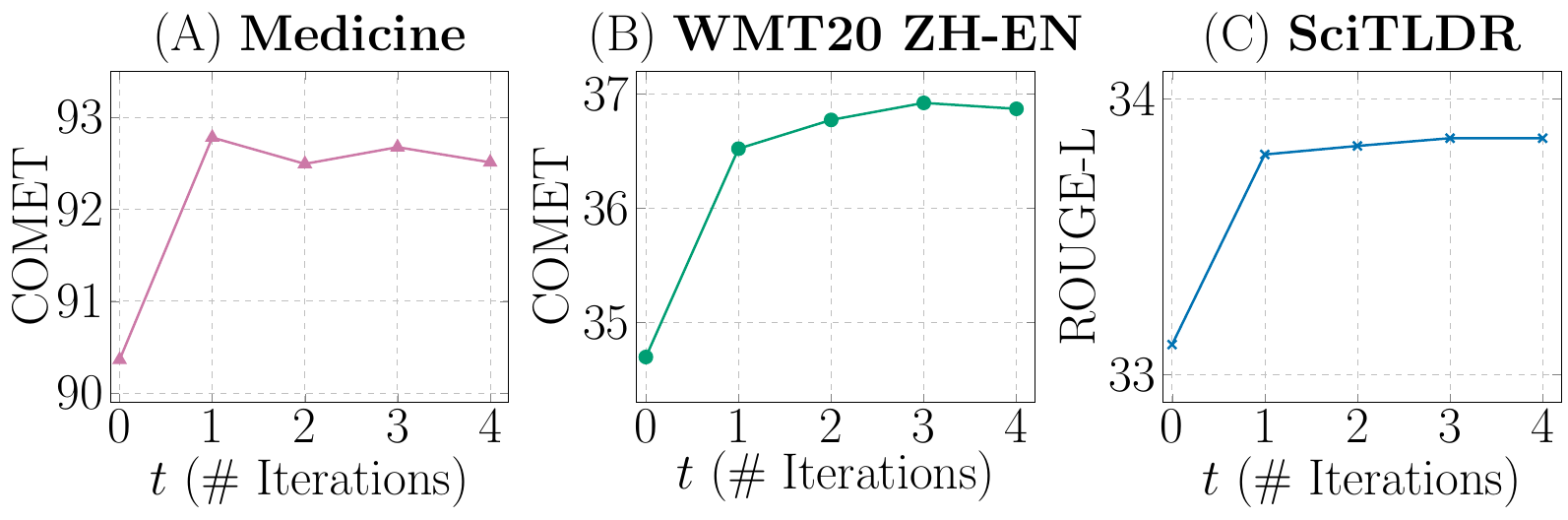}
\caption{Effects of iterations on dev.\ performance. Iteration 0 refers to  the initial decoding from $f$. Every iteration consists of $g$'s decoding with $f$'s guidance followed by $f$'s decoding with $g$'s guidance.
The values of $\lambda$s are kept the same over all iterations for simplicity. Initially, we explored gradually increasing the $\lambda$s as f and g’s outputs become closer, but we found no substantial performance gain.
}
\label{fig:iterations}
\end{figure}

\paragraph{Iterations.}
So far, we have only applied one iteration of \twist decoding, but Fig.\ \ref{fig:iterations} plots performance over multiple iterations.
Iteration 0 signifies $f$'s initial decoding ($\vy^{(0)}$ in Fig.\ \ref{fig:twist_decoding}), and every iteration involves $g$'s decoding with $f$'s guidance ($\vz^{(t)}$) and its reverse ($\vy^{(t)}$).
We observe that the first iteration brings most of the performance gains.
This makes \twist decoding practically appealing, as it improves performance without much increase in the computation or inference time (see below).

\paragraph{Inference Time.}
Table \ref{tab:inference_time} reports the runtime of each decoding method, relative to $f$'s decoding in isolation.
We use batch size 1 on the same single A100-SXM GPU and measure the wall-clock time from when all models are loaded until all outputs are obtained.
As expected, \twist decoding results in a slowdown compared to decoding in isolation, but the increase in time is only 50\%.
The inference time for \twist decoding is much shorter than the sum of $f$ and $g$ in isolation (1.4$\times$ vs.\ 2.1$\times$ on medical translation) because 1) the encoder computation for $f$ and $g$ can be parallelized and 2) the encoder computation for $f$ is done only once while we need two runs of $f$'s decoder.
We leave it to future work to further speed up \twist decoding; since the slowdown of \twist decoding primarily comes from the decoder, it can be sped up by best-first beam search \cite{meister-etal-2020-best}, a deep-encoder, shallow-decoder strategy \cite{deepshallow}, or a fast, linear-complexity variant of the transformer decoder \cite{peng2021rfa,kasai2021t2r} that is shown to retain the performance of the standard encoder-decoder transformer.
Another approach could be sequence-level knowledge distillation \cite{Kim2016SequenceLevelKD}, which has proven successful in speeding up an ensemble translation model \cite{ensemble-distillation}.

\begin{table}[h]
 \addtolength{\tabcolsep}{-3.5pt}  
\centering
\small
\begin{tabular}{@{} l ccc m{0.005em} ccc  @{}}
\toprule[.1em]

\textbf{Inference}
& \multicolumn{3}{c}{\textbf{Medicine}}
&& \multicolumn{3}{c}{\textbf{WMT20 ZH$\rightarrow$EN}}\\
\cmidrule(lr){2-4}
\cmidrule(lr){6-8}
\textbf{Method}
& $f$ & $g$
& Time && $f$ & $g$ & Time
\\

\midrule[.1em]
Isolation
&  Domain
& --
& 1.0$\times$ 

&& R2L
& --
& 1.0$\times$
\\

Isolation
&  Generic
& --
& 1.1$\times$

&& L2R
& --
& 1.0$\times$
\\

Rerank
&  Domain
& Generic
& 1.0$\times$

&& R2L
& L2R
& 1.0$\times$
\\

\twist
&  Domain
& Generic
& 1.4$\times$

&& R2L
& L2R
& 1.5$\times$
\\

\bottomrule[.1em]
\end{tabular}
\caption{
Inference time relative to a single model decoded in isolation. It is measured on the same single Nvidia A100-SXM GPU with batch size 1.
We measure the wall-clock time from when the models are loaded until the
last sentence is translated on the test data.
} 
\label{tab:inference_time}
\end{table}

\begin{figure}[h!]
\centering
\includegraphics[width=0.23\textwidth]{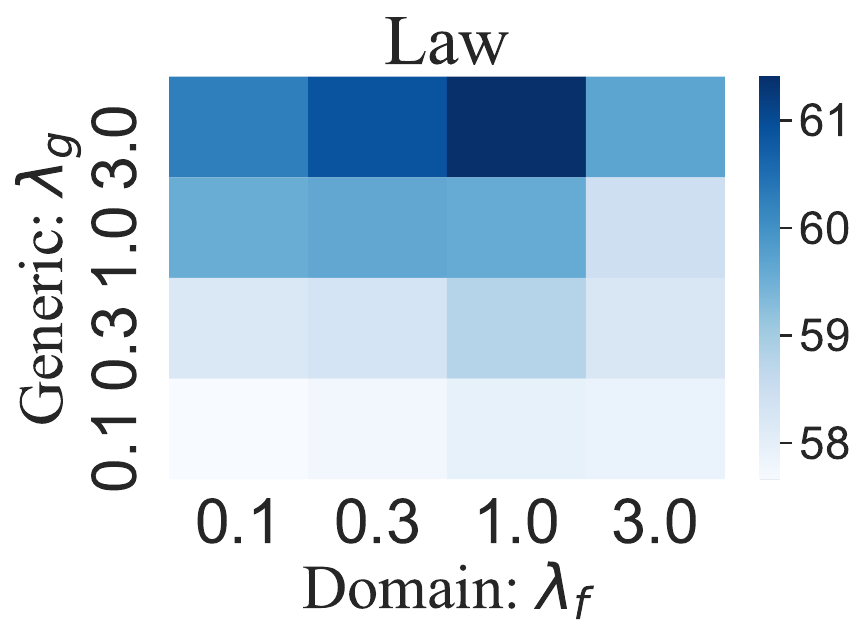}
\hspace{-0.1cm}
\includegraphics[width=0.23\textwidth]{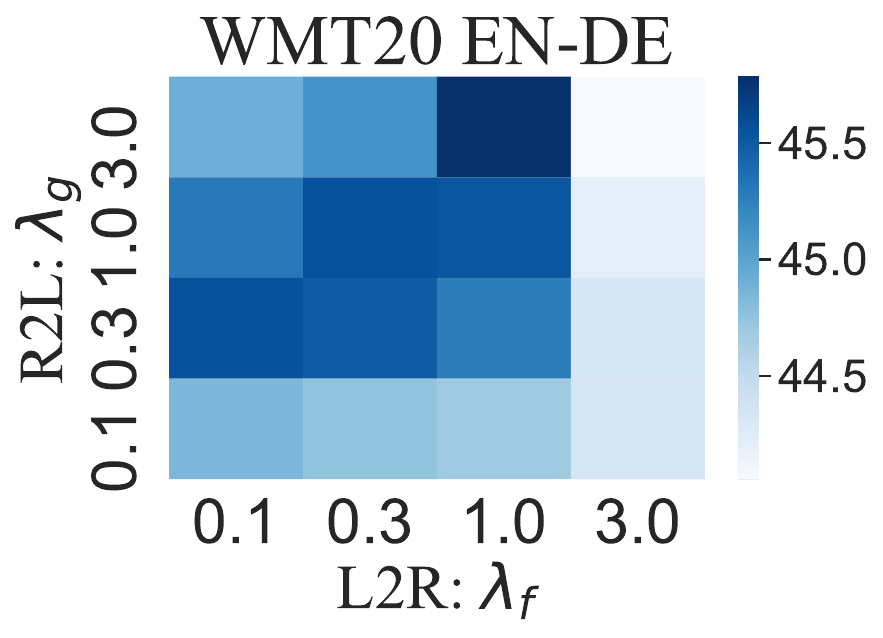}
\caption{Dev.\ set performance measured in the COMET score \cite{rei-etal-2020-comet,comet-wmt} with varying $\lambda_f$ and $\lambda_g$. See Appendix \S\ref{appendix_sec:heatmap} for other configurations.}
\label{fig:heatmap}
\end{figure}

\begin{table}[h!]
 \addtolength{\tabcolsep}{-3.5pt}  
\centering
\small
\begin{tabular}{@{} l cc m{0.0001em} cc  @{}}
\toprule[.1em]

\textbf{Dev.\ Set Results}
& \multicolumn{2}{c}{\textbf{Medicine}}
&& \multicolumn{2}{c}{\textbf{WMT20 ZH-EN}}\\
\cmidrule(lr){2-3}
\cmidrule(lr){5-6}
\textbf{Distance Function}
& \textsc{COMET} & \textsc{BLEU}
&& \textsc{COMET} & \textsc{BLEU}
\\

\midrule[.1em]
Original
&  92.8
& 58.1
&&  \textbf{36.5}
& \textbf{26.4}
\\

One Candidate
& \textbf{93.2}
& \textbf{58.2}
&&  35.4
&  25.8
\\

Embed.\ Distance
& 92.8
& 57.9
&& \textbf{36.5}
& 26.2

\\

\bottomrule[.1em]
\end{tabular}
\caption{
Variants of the distance function in \twist decoding.
$f$ is the domain model and $g$ is the generic model for medical translation (German$\rightarrow$English). $f$ is R2L and $g$ is L2R for WMT20 Chinese$\rightarrow$English.
} 
\label{tab:variants}
\end{table}

\newcolumntype{?}{!{\vrule width .1em}}

\begin{table*}[h!]
 \addtolength{\tabcolsep}{-3.5pt}  
\centering
\small
\begin{tabular}{@{} L{1.7cm} L{7.1cm} ? L{6.7cm} @{}}
\toprule[.1em]
\textbf{Medicine}
&
Domain (Isolation) \thumbup \quad Generic (Isolation) \thumbdown
&
Domain (Isolation) \thumbdown \quad Generic (Isolation) \thumbup 
\\
\midrule[.05em]
\textbf{Reference}
& 
If signs and symptoms of tardive dyskinesia appear in a patient on ABILIFY, dose reduction or discontinuation should be considered. 
&
In placebo-controlled trials, the incidence of akathisia in bipolar patients was 12.1\% with aripiprazole and 3.2\% with placebo.
\\
\midrule[.05em]
\textbf{Domain}
&
If signs and symptoms of \greenbox{tardive dyskinesia} appear in one patient on ABILIFY, a dose reduction or discontinuation should be considered.
&
In placebo-controlled \greenbox{trials}, the incidence of akathisia in \redbox{bipolar disorder} was \redbox{12.1\% and 3.2\% with aripiprazole}.
\\

\midrule[.05em]

\textbf{Generic}
&
If a patient treated with ABILIFY shows signs and symptoms of \redbox{late dyskinesia}, it should be considered to reduce the dose or stop treatment.
&
In placebo-controlled \redbox{studies}, the incidence of akathisia in \greenbox{bipolar patients} was \greenbox{12.1\% with aripiprazole and 3.2\% with placebo}.
\\

\midrule[.05em]
\textbf{\twist}
\quad\quad $f$: Domain
\quad\quad $g$: Generic
& 
If signs and symptoms of \greenbox{tardive dyskinesia} appear in one patient on ABILIFY, a dose reduction or discontinuation should be considered.
&
In placebo-controlled \greenbox{trials}, the incidence of akathisia in \greenbox{bipolar patients} was \greenbox{12.1\% with aripiprazole and 3.2\% with placebo}.
\\
\bottomrule[.1em]
\end{tabular}
\caption{
Example outputs from machine translation on the medical domain.
For \twist decoding, $f$ is the domain model, and $g$ is the generic model.
In the left section, the generic model fails to capture technical terminology (\redbox{late dyskinesia} vs.\ \greenbox{tardive dyskinesia} for the German term, \textit{Sp\"atdyskinesie}), and \twist decoding chooses the correct term of \greenbox{tardive dyskinesia} from the domain model.
In the right example, the domain model has a problem in coordination (\redbox{12.1\% and 3.2\% with aripirazole} vs.\ \greenbox{12.1\% with aripirazole and 3.2\% with placebo}), and \twist decoding successfully benefits from the accurate translation of the generic model.
} 
\label{tab:examples}
\end{table*}

\paragraph{Sensitivity Analysis on Distance Coefficients.}
As discussed in \S\ref{sec:mutual_guidance}, $\lambda_f$ and $\lambda_g$ weight the distance terms \textit{from} $f$ and $g$ respectively.
We tuned $\lambda_f$ and $\lambda_g$ on the dev.\ set from the range of $\{0.1, 0.3, 1.0, 3.0\}$.
Fig.\ \ref{fig:heatmap} visualizes how they affect the overall performance on the dev.\ sets.
$\lambda_g\!>\!\lambda_f$ generally yields good performance, suggesting the effectiveness of the initial exploration by $g$ with relatively weaker guidance from $f$.

\paragraph{Variants of Distance Functions.}
We experiment with two variants of distance terms (Table \ref{tab:variants}): 1) \textit{one candidate}, which measures the distance from the 1-best candidate from the other model (vs.\ minimization over multiple candidates; \S\ref{sec:mutual_guidance}) and 2) \textit{embed.\ distance}, which calculates the distance based on the Euclidean distance between the embeddings.
Here the embeddings are taken from the output layer of the decoder.
Overall, both variants yield similar performance to the original distance function, but the \textit{one candidate} method has a substantial performance drop on WMT20 ZH-EN.
Note also that the \textit{embed.\ distance} method necessitates additional distance computations between the token embeddings.
This result illustrates that our original distance function is a simple yet effective design choice.

\paragraph{Examples.}
Seen in Table \ref{tab:examples} are example German$\rightarrow$English translations from the medical domain.
The left section presents a case where the domain model translates the technical term, \emph{Sp\"atdyskinesie}, into the corresponding English term: \greenbox{tardive dyskinesia}.
The generic model, on the other hand, generates a literal translation: \redbox{late dyskinesia}. In the right section, the domain model fails to handle the coordinate structure: \redbox{12.1\% and 3.2\% with aripiprazole} vs.\ \greenbox{12.1\% with aripiprazole and 3.2\% with placebo}.
Further, the final output has wording closer to the reference translation: \greenbox{trials} vs.\ \redbox{studies} and \greenbox{bipolar patients} vs.\ \redbox{bipolar disorder}.
These examples illustrate that \twist decoding benefits from the complementary strengths of the domain and generic models.

%% file: text/related.tex
\section{Further Related Work}
\paragraph{Decoding from Multiple Models.}
Much early work proposed methods to generate text from multiple models especially for machine translation (often called \textit{consensus-based decoding}; \citealp{bangalore2001,bangalore-etal-2002-bootstrapping,matusov-etal-2006-computing,rosti-etal-2007-combining,SimBGSW07,hildebrand-vogel-2008-combination}).
Most of these methods limit their search space to n-best candidates from individual translation models \cite{li-etal-2009-collaborative}, contrasting with our \twist decoding where one model can update its translation outputs under the guidance of another model.
\textit{Collaborative decoding} \cite{li-etal-2009-collaborative} trains a separate feature-based scorer that measures the consensus between phrase-based Chinese-to-English translation models.
Several recent works proposed inference algorithms for decoding from multiple generators for specific tasks, such as detoxification and abductive reasoning \cite{west-etal-2021-reflective,liu-etal-2021-dexperts}.

\paragraph{Alternatives to Left-to-Right Decoding.}
We showed that \twist decoding can be used to benefit from models with diverging generation order. 
Several prior works proposed approaches for generating text in a different fashion than the standard left-to-right order.
For example, much recent work explored non-autoregressive generation \interalia{Gu2017NonAutoregressiveNM,Lee2018DeterministicNN,Mansimov2019AGF,Ghazvininejad2019MaskPredictPD,Kasai2020ParallelMT} primarily to parallelize and speed up inference. 
More specifically, several works introduced training and/or inference algorithms that combine left-to-right and right-to-left models for machine translation \cite{zhou-etal-2019-synchronous} and commonsense inference \cite{dynamic_recurrent}. 
\citet{qin-etal-2020-back} incorporated right (future) context into a left-to-right language model by iterative gradient-based updates on the output representations.
Those algorithms are designed specifically for the combination of left-to-right and right-to-left generation and cannot be easily extended to more general situations, such as diverging tokenization and vocabularies where \twist decoding has been shown effective.

%% file: text/conclusion.tex
\section{Conclusion}\label{sec:conclusion}
We presented \twist decoding, a general inference algorithm that generates text from diverse models without the assumption of a shared vocabulary, tokenization, or generation order.
Our method enables diverse models to guide each other, thereby outperforming individual models over various scenarios, even when one of the models is much weaker because of limited data.
We also demonstrated that \twist decoding can be viewed as a generalization and improvement of the commonly-adopted reranking heuristic.
As it only requires a small change in code, we hope that researchers and practitioners will explore complementary strengths of diverse generation models through \twist decoding.

%% file: text/limitation.tex
\section*{Limitations}
We evaluated our decoding method that combines generation models both on machine translation and scientific paper summarization over several scenarios: combining 1) generic and domain-specific models, 2) left-to-right and right-to-left generation models, and 3) models that generate using different conditioning inputs.
Our machine translation experiments span diverse domains, including medical and legal text.
We also presented results from recent English-to-German and Chinese-to-English WMT data.
Nonetheless, our domain translation experiments are limited to German-to-English, and we only dealt with scientific papers written in English, mainly due to availability of data.
There are also many other language generation tasks for which our method can be useful.
Since we open-source our codebase built on top of a popular library, we hope that practitioners will use it for applications of their interest and further assess our decoding algorithm in many application scenarios.

Evaluating language generation remains a challenging research problem.
We carefully set up our experiments to mitigate potential evaluation issues.
The WMT 2020 test data consist only of news text written in the original language, in contrast to the test data from WMT 2018 \cite{bojar-etal-2018-findings} or earlier.
The WMT 2020 EN$\rightarrow$DE and DE$\rightarrow$EN test data that we used thus come from completely different documents.
This avoids the translationese effect that would overestimate the translation performance due to the simplicity of translated text \cite{Graham2020TranslationeseIM}.
Moreover, the WMT 2020 test data for English-to-German and Chinese-to-English translation have multiple reference translations per instance, which increases the correlation of reference-based, automatic evaluations with human judgment \cite{billboard}.
We presented results using automatic metrics from recent work \cite{comet-wmt} as well as conventional, n-gram overlap metrics \cite{Papineni2001BleuAM,Lin2004ROUGEAP}.
Recent automatic metrics have shown to have higher correlation with human judgments, but human judgments are sometimes inconsistent, especially when crowdsourced \cite{clark-etal-2021-thats,kasai2021thumb}.
Since our decoding method is a simple modification of the widely-used beam search algorithm, we hope that it will be tested and used in real-world systems of language generation.

%% file: text/acknowledgement.tex
\section*{Acknowledgements}
We thank Hila Gonen, Phillip Keung, the ARK group at the UW, and the Mosaic team at the Allen Institute for AI for their helpful feedback on this work.
This work was supported in part by the DARPA MCS program through NIWC Pacific (N66001-19-2-4031) and Google Cloud Compute. 
Hao Peng was supported by a Google Ph.D.\ Fellowship.

%% file: text/appendix.tex
\begin{appendices}

\section{Hyperparameters and Settings}
We provide training and implementation details for easy replication of our work.
\subsection{Domain Machine Translation}
\label{appendix:domain_mt}
We generally follow the preprocessing and subword tokenization from \citet{koehn-knowles-2017-six,hu-etal-2019-domain}.
Table \ref{tab:base-setting} lists the hyperparameters and setting on \texttt{fairseq} that we use for all domain-specific translation models.
All embeddings are shared \cite{Press2017UsingTO,Inan2017TyingWV}.
We choose the checkpoint that achieved the best loss on the validation data.

\begin{table}[h]
\small
\centering
\begin{tabular}{@{} l@{\hspace{-0.2cm}} r @{}}
\toprule[.1em]
\textbf{Hyperparameter} & \textbf{Value}\\
\midrule[.1em]
label smoothing & 0.1\\
\# max tokens & 8192\\
dropout rate & 0.1 \\
encoder embedding dim  & 512\\
encoder ffn dim  & 2048\\
\# encoder attn heads & 8\\
decoder embedding dim  & 512\\
decoder ffn dim  & 2048\\
\# decoder attn heads & 8\\
max source positions & 1024 \\
max target positions & 1024 \\
Adam lrate& $5\times 10^{-4}$ \\
Adam $\beta_1$& 0.9\\
Adam $\beta_2$& 0.98\\
lr-scheduler &  inverse square \\
warm-up lr & $1\times 10^{-7}$ \\
\# warmup updates & 4000 \\
\# max updates &  600K \\
\# GPUs &  8 \\
length penalty & 0.6\\
\bottomrule[.1em]
\end{tabular}
\caption{Domain translation \texttt{fairseq} hyperparameters and setting. We generally follow the base-sized configuration from \citet{Vaswani2017AttentionIA}.}
\label{tab:base-setting}
\end{table}

\subsection{Left-to-Right and Right-to-Left}
\label{appendix:l2r_r2l}
\paragraph{WMT20 ZH-EN}
Table \ref{tab:large-setting} lists the hyperprameters and setting on \texttt{fairseq} that we use for left-to-right and right-to-left models on the WMT20 ZH-EN dataset.
We generally follow the preprocessing and tokenization from \citet{billboard}.
We use \texttt{newstest-2019} as the dev.\ set and the official training data.\footnote{\url{http://www.statmt.org/wmt20/translation-task.html}.}
We apply Moses tokenization \cite{koehn-etal-2007-moses} and BPE with 32K operations \cite{sennrich-etal-2016-neural} to English text. 
We tokenize Chinese text with the Jieba package,\footnote{\url{https://github.com/fxsjy/jieba}.} following \citet{Hassan2018AchievingHP}.
Separately from English, BPE with 32K operations is then applied to Chinese.
The decoder input and output embeddings are tied.

\paragraph{WMT20 EN-DE}
The same hyperparameters are chosen as in WMT20 ZH-EN (Table \ref{tab:large-setting}). 
We again follow \citet{billboard} and preprocess both English and German text by the Moses tokenizer and \textit{joint} BPE with 32K operations. All embeddings are shared.

\begin{table}[h]
\small
\centering
\begin{tabular}{@{} l@{\hspace{-0.2cm}} r @{}}
\toprule[.1em]
\textbf{Hyperparameter} & \textbf{Value}\\
\midrule[.1em]
label smoothing & 0.1\\
\# max tokens & 4096 \\
dropout rate & 0.1\\
encoder embedding dim  & 1024\\
encoder ffn dim  & 4096\\
\# encoder attn heads & 16\\
decoder embedding dim  & 1024\\
decoder ffn dim  & 4096\\
\# decoder attn heads & 16\\
max source positions & 1024 \\
max target positions & 1024 \\
Adam lrate& $5\times 10^{-4}$ \\
Adam $\beta_1$& 0.9\\
Adam $\beta_2$& 0.98\\
lr-scheduler &  inverse square \\
warm-up lr & $1\times 10^{-7}$ \\
\# warmup updates & 4000 \\
\# max updates &  600K \\
\# GPUs &  8 \\
length penalty & 0.6\\
\bottomrule[.1em]
\end{tabular}
\caption{
L2R and R2L translation \texttt{fairseq} hyperparameters and setting. We generally follow the large-sized configuration from \citet{Vaswani2017AttentionIA}.
}
\label{tab:large-setting}
\end{table}

\subsection{SciTLDR}
\label{appendix:tldr}
We use two BART-based pretrained models from \citet{cachola-etal-2020-tldr}: the abstract-only version of BART and the AIC version of \textsc{Catts}$_\mathrm{XSUM}$.\footnote{They are both available at \url{https://github.com/allenai/scitldr}.}
These two models are both BART-based models; \textsc{Catts}$_\mathrm{XSUM}$ is obtained by finetuning BART on the XSUM dataset \cite{xsum2018} with multitask scaffolding \cite{cachola-etal-2020-tldr}.

\begin{table}[h]
 \addtolength{\tabcolsep}{0.0pt}  
\centering
\small
\begin{tabular}{@{} l cc  m{0.5cm} cc  @{}}
\toprule[.1em]

&&&&\multicolumn{2}{c}{\textbf{Tuned $\lambda$}}
\\
\cmidrule(lr){5-6}
\textbf{Dataset}
&
$f$
& $g$
&& $\lambda_f$
& $\lambda_g$
\\

\midrule[.1em]

\multirow{2}{*}{Medicine}
& Domain
& Generic
&& 0.1
& 0.3
\\

& Generic
& Domain
&& 0.1
& 3.0
\\

\midrule[.05em]

\multirow{2}{*}{Law}
& Domain
& Generic
&& 1.0 
& 0.1 
\\
& Generic
& Domain
&& 0.1
& 3.0
\\

\midrule[.05em]

\multirow{2}{*}{Koran}
& Domain
& Generic
&& 1.0 
& 3.0
\\
& Generic
& Domain
&& 0.3
& 3.0
\\

\midrule[.05em]

\multirow{2}{*}{Subtitles}
& Domain
& Generic
&& 1.0
& 1.0
\\
& Generic
& Domain
&& 1.0 
& 1.0
\\

\midrule[.05em]

\multirow{2}{*}{WMT20 ZH-EN}
& L2R
& R2L
&& 1.0
& 3.0
\\
& R2L
& L2R
&& 0.1 
& 3.0
\\

\midrule[.05em]

\multirow{2}{*}{WMT20 EN-DE}
& L2R
& R2L
&& 0.3 
& 0.3
\\
& R2L
& L2R
&& 0.1 
& 0.3
\\

\midrule[.05em]

\multirow{2}{*}{SciTLDR}
& Abst.\
& AIC
&& 1.0
& 3.0
\\
& AIC
& Abst.\
&& 0.3
& 3.0
\\

\bottomrule[.1em]
\end{tabular}
\caption{
Selected $\lambda_f$ and $\lambda_g$ values.
} 
\label{tab:lambda_tuning}
\end{table}

\subsection{$\lambda$ Tuning}
We tune $\lambda_f$ and $\lambda_g$ from $\{0.1, 0.3, 1.0, 3.0\}$, based on the dev.\ BLEU/ROUGE-L score on machine translation and paper summarization, respectively.
Table \ref{tab:lambda_tuning} reports the selected $\lambda$ values in all scenarios.

\begin{figure*}[h!]
\centering
\includegraphics[width=0.24\textwidth]{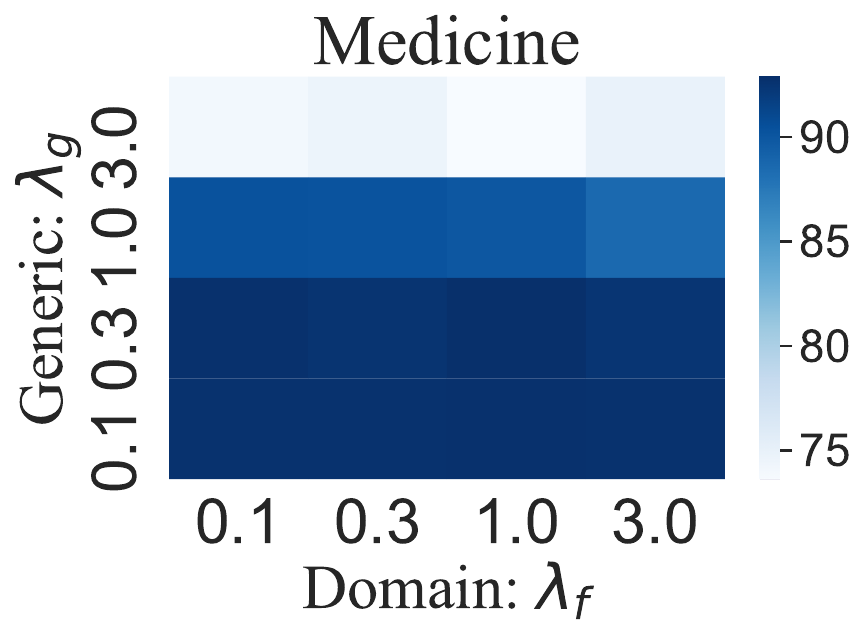}
\hspace{-0.1cm}
\includegraphics[width=0.24\textwidth]{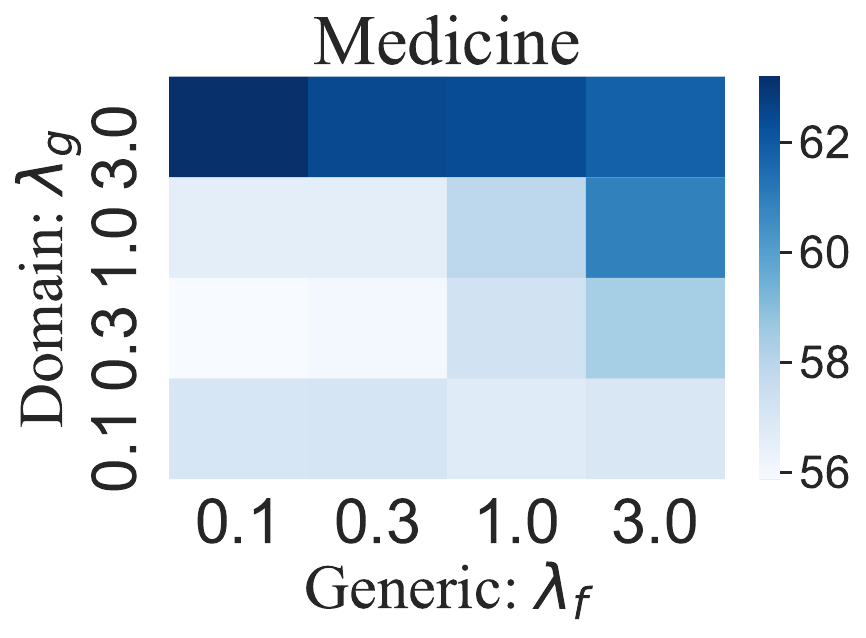}
\hspace{-0.1cm}
\includegraphics[width=0.24\textwidth]{figs/heatmaps/acquis_domain.pdf}
\hspace{-0.1cm}
\includegraphics[width=0.24\textwidth]{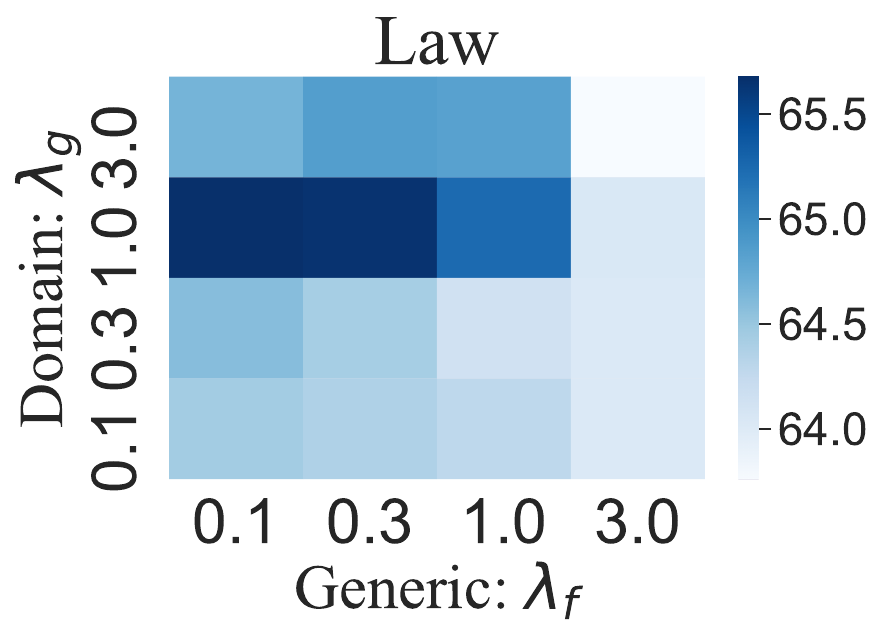}

\includegraphics[width=0.24\textwidth]{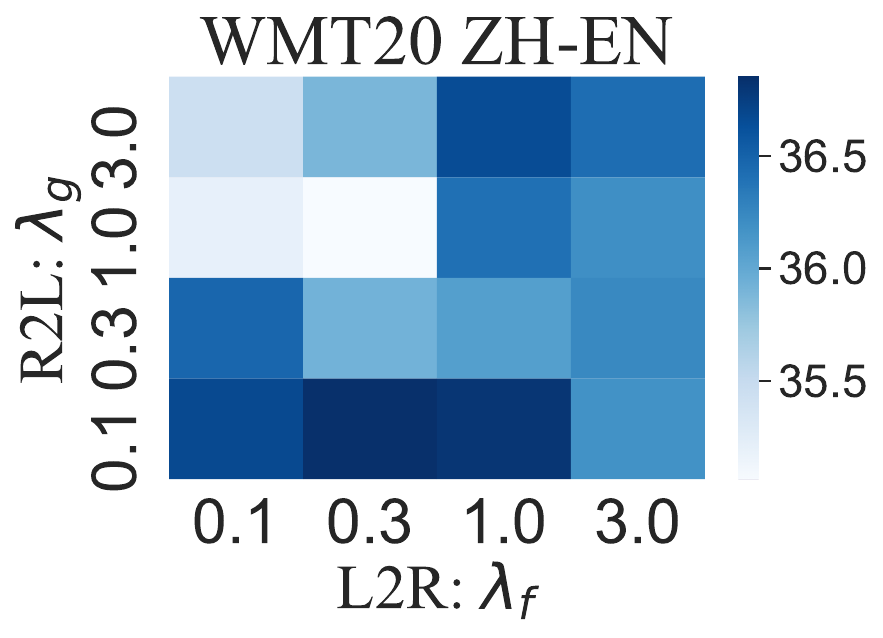}
\hspace{-0.1cm}
\includegraphics[width=0.24\textwidth]{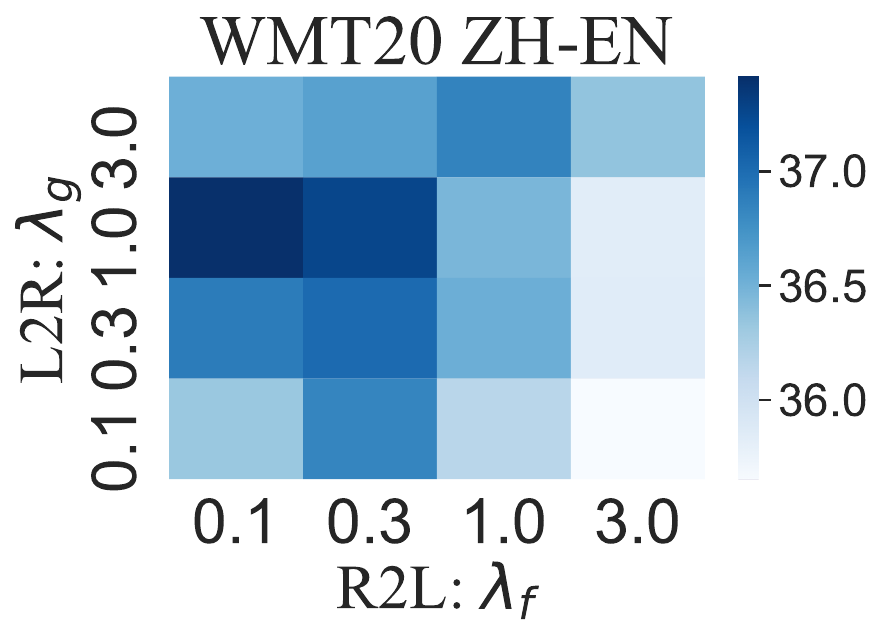}
\hspace{-0.1cm}
\includegraphics[width=0.24\textwidth]{figs/heatmaps/en-de_l2r.pdf}
\hspace{-0.1cm}
\includegraphics[width=0.24\textwidth]{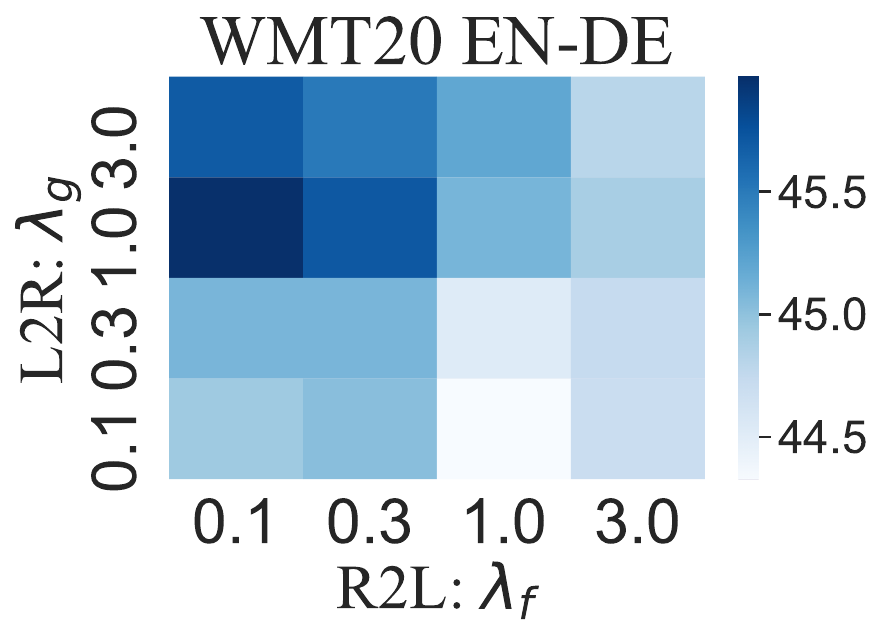}

\caption{Dev.\ set performance measured in the COMET score \cite{rei-etal-2020-comet,comet-wmt} with varying $\lambda_f$ and $\lambda_g$.}
\label{appendix_fig:heatmap}
\vspace{-0.0cm}
\end{figure*}

\section{Sensitivity Analysis on $\lambda$}
\label{appendix_sec:heatmap}
Fig.\ \ref{appendix_fig:heatmap} presents the sensitivity analysis in the COMET score over many scenarios. 
Apart from a few exceptions, $\lambda_g > \lambda_f$ tends to yield good performance, suggesting the effectiveness of the initial exploration by $g$ with relatively weaker guidance from $f$.
\end{appendices}